# Comparison of decision trees with Local Interpretable Model-Agnostic Explanations (LIME) technique and multi-linear regression for explaining support vector regression model in terms of root mean square error (RMSE) values.

**Amit Thombre**

(amit_thombre@hotmail.com)

In this work the decision trees are used for explanation of support vector regression model. The decision trees act as a global technique as well as a local technique. They are compared against the popular technique of LIME which is a local explanatory technique and with multi linear regression. It is observed that decision trees give a lower RMSE value when fitted to support vector regression as compared to LIME in 87% of the runs over 5 datasets. The comparison of results is statistically significant. Multi linear regression also gives a lower RMSE value when fitted to support vector regression model as compared to LIME in 73% of the runs over 5 datasets but the comparison of results is not statistically significant. Also, when used as a local explanatory technique, decision trees give better performance than LIME and the comparison of results is statistically significant.

1. Introduction

In all domains where machine learning models are used it has become necessary to understand the decision-making process and what contributes to a particular prediction. Generally, for achieving higher accuracy complex or black box machine learning models are getting used. These models are not self-explanatory and lack transparency. Support vector regression is one such black box technique which is widely used because it can handle non-linear data by the usage of kernels. To explain such models, interpretable models such as multi linear regression and decision trees are used among various other techniques. Decision trees are considered highly interpretable machine learning models due to their transparent and easily understandable decision-making process. LIME is another technique which is popularly used for local explanations. It is a technique that approximates any black box machine learning model with a local, interpretable model to explain each individual prediction.

The interpretability of models can either be a global level or on a local level. The interpretability can either be on a global level, meaning it makes different models comparable with each other, by summarizing their performance on the entire data. The local level interpretability gives insight into how a classification or regression in the case of a single prediction is made. Most of the work involves local interpretability [26,27, 36, 14, 33, 32, 16, 24, 31, 35, 22, 23, 5, 7, 19, 15, 1, 13, 34, 2, 8, 3,4]. The global comparison work can be found in [36, 14, 33, 24, 23, 10, 13, 4, 18, 3, 21].

The literature is missing a detail study on decision trees as an explanatory technique for support vector regression models and its comparison with LIME and multi linear regression in terms of RMSE. The hypothesis for this study is that decision trees will provide more accurate explanation compared to LIME in terms of RMSE values. This may be because decision trees are capable of capturing non-linear relationships and thus will better align with the underlying structure of SVR predictions. On the other hand, LIME constructs a locally interpretable model such as a linear model in the vicinity of a specific instance to approximate the behaviour of the complex model. In doing so the approximation might not fully capture the non-linear intricacies of the original model. This may lead to greater values of RMSE of explanations. This paper proves the above hypothesis by carrying out experiments on 5 datasets. This paper will try to answer the question as to which is the best explanatory technique for support vector regression in terms of RMSE.

The paper describes Decision trees and LIME in section 2. Section 3 gives the data used in this study. Section 4 lists down the experiments carried out. Section 5 is the results and discussion section.

Section 6 gives the conclusion of the paper. The paper ends with future work in section 7 and references in section 8.

2. **Decision trees and LIME**

   Decision trees are considered to be highly interpretable machine learning models. This is due to the following reasons: -

   a. Tree Structure
   The visual representation of a decision tree is represented by branches and nodes in a hierarchical structure. It is easy to follow the logic as each node represents a decision based on a specific feature.
   b. Human-Readable rules
   By traversing the tree from top to bottom or from root to leaf, one can understand how a prediction is made by forming the if-else rules. These rules are human-readable.
   c. Feature importance
   The features which are closer to the root are generally more influential in making decisions and thus. give the feature importance.
   d. Explanatory Power
   The reasoning behind each decision is explicitly shown by the decision trees. By following the branches of the tree, one can understand how a decision is arrived at.
   e. Visual Representation
   The decision trees can be easily represented as diagrams with root, leaves and the rules at each node.
   f. Handling Non-linearity
   The decision trees can capture the non-linear relationships between the features and the target variable enhancing interpretability in non-linear scenarios.

   LIME (Local Interpretable Model-agnostic Explanations) is a technique used to explain complex machine learning model prediction locally. The following reasons explain why LIME is used as a popular interpretation technique: -

   a. Local Interpretability
   LIME provides explanations of the predictions of a black box model at a specific instance or in the vicinity of a specific instance. LIME creates a simple interpretable model that approximates the behavior of the black-box model locally.
   b. Model Agnosticism
   LIME is model agnostic meaning it can be applied to explain any black box model such as neural networks, support vector machines or ensemble methods. This flexibility allows LIME to be sued for any machine learning algorithm without knowing its internal workings.
   c. Visual Explanations
   LIME can provide visual explanations of its local interpretations making it easier for users to understand and interpret the behavior of the black-box model in specific instances. Visualizations such as feature importance plots aid in interpreting the model's predictions.
   d. Domain-Agnostic
   LIME can be applied across various domains and types of data, making it versatile interpretation technique.

3. **Data**
   The data for regression was obtained from UCI machine learning repository. The details of the data are given below.

| Name of the dataset | Number of columns | Number of rows. |
|---|---|---|
| Wine | 12 | 1359 |
| /Boston Housing | 14 | 506 |
| Yatch hydrodynamics | 7 | 308 |
| Computer Hardware | 10 | 209 |
| Auto | 7 | 398 |

Table 1. Details of the datasets used.

4. **Experiments**
   1. In all 15 runs were conducted on 5 datasets thus having 3 runs for each dataset.
   2. Initially multi-linear regression and support vector regression was carried out on each dataset using random number of features in each run. The number of features were varied to establish that the results are independent of the features used. This was for evaluating which regression was better on the data.
   3. Then LIME was used to explain support vector regression model.
   4. Then each run was carried using decision trees and multi linear regression to explain support vector regression.
   5. All of the code was developed using R statistical software. Statistical test of paired Wilcoxon was used to compare the results of any 2 techniques.

5. **Results and Discussion**
   1. In all runs the multi-linear regression RMSE was greater than the support vector regression RMSE. This ensured that support vector regression was a better model for the dataset at hand and a black box model gave a better performance than an explanatory technique. So, all of the cases were genuine for them to be explained by a simpler and interpretable technique.
   2. The above table shows the comparison of accuracies obtained by using multi-linear regression, decision trees and LIME over various variables on 5 datasets. The table shows that for 87% of the runs the RMSE of decision trees is less than that of LIME and for 73% of the runs the RMSE of multi-linear regression is less than that of LIME.
   3. The statistical comparison of RMSE values of decision trees with LIME using paired Wilcoxon test gives the p-value as 0.022 which is less than the significance level alpha of 0.05 and thus the results are statistically significant.
   4. The statistical comparison of RMSE values of multi-linear regression with LIME using paired Wilcoxon test gives the p-value as 0.252 which is greater than the significance level alpha of 0.05 and thus the results are not statistically significant.
   5. The statistical comparison of RMSE values of decision tress with multi-linear regression comes out to be not significant. In 60% of the runs, decision tree RMSE values are less than that of multi-linear regression RMSE values.
   6. The local prediction squared error values using decision trees were compared against those of LIME and the total count of each technique superseding the other technique over all records of the test dataset for a run was calculated as a percentage. The p-value of comparison of the percentage comparison shows that the results were statistically significant. The following table shows the results.
   7. Similarly, as above the percentage comparison of records of local prediction squared error values of multi-linear regression with LIME turns out to be also statistically significant.
   8. In 67% of the runs the percentage of local prediction squared error values of decision trees was greater than that of multi-linear regression but the comparison was not statistically significant.

| Dataset | Number of variables used | RMSE using multi-linear regression | RMSE using Decision trees | RMSE using LIME |
|---|---|---|---|---|
| Wine | 9 | 0.065 | 0.145 | 0.256 |
| Wine | 8 | 0.132 | 0.165 | 0.248 |
| Wine | 6 | 0.176 | 0.193 | 0.257 |
| Boston Housing | 10 | 3.964 | 3.01 | 6.8 |
| Boston Housing | 11 | 3.597 | 3.183 | 4.41 |
| Boston Housing | 12 | 3.464 | 2.55 | 4.24 |
| Yatch hydrodynamics | 3 | 9.139 | 1.7 | 7.31 |
| Yatch hydrodynamics | 5 | 8.994 | 1.883 | 7.497 |
| Yatch hydrodynamics | 2 | 8.63 | 1.39 | 5.32 |
| Computer Hardware | 5 | 26.73 | 61.26 | 129.82 |
| Computer Hardware | 6 | 0.012 | 6.89 | 18.43 |
| Computer Hardware | 3 | 42 | 70.56 | 48.1 |
| Auto | 6 | 2.2 | 1.69 | 2.52 |
| Auto | 4 | 2.43 | 1.83 | 2.88 |
| Auto | 5 | 3.34 | 3.227 | 2.05 |

Table 2. Comparison of accuracies using decision trees, multi-linear regression and LIME over 5 datasets.

**6. Conclusion**
1. The results from Table 2 and the statistical significance of comparison of results show that the decision trees explain support vector regression model in a better way than LIME when used as a global technique.
2. The results from Table 3 and the statistical significance of comparison of results show that decision trees are also better in performance than LIME when it comes to explaining local prediction values.
3. The results from table 2 and the statistical significance of comparison of results show that multi-linear regression is also better in performance than LIME when it comes to explaining local prediction values.
4. This exceptional ability of decision trees to explain support vector regression model globally as well as locally in a better way than multi-linear regression and LIME is due to its ability to handle non-linear data in a better way. LIME fits a linear model to the perturbed instances of the selected instance. In doing so, LIME is missing the non-linear intricacies of the original model and multi-linear regression cannot handle non-linear data. This leads to greater values of RMSE of explanations.

| Dataset | Number of test dataset records over which decision trees squared error value is less than of LIME(x1) | Number of test dataset records over which multi-linear regression squared error value is less than of LIME(x2) | Total number of test dataset records(T) | x1/T*100 | x2/T*100 |
|---|---|---|---|---|---|
| Wine | 236 | 260 | 272 | 86.76 | 95.59 |
| Wine | 231 | 241 | 272 | 84.93 | 88.60 |
| Wine | 226 | 230 | 272 | 83.09 | 84.56 |
| Boston Housing | 71 | 71 | 102 | 69.61 | 69.61 |
| Boston Housing | 67 | 66 | 102 | 65.69 | 64.71 |
| Boston Housing | 75 | 56 | 102 | 73.53 | 54.90 |
| Yatch hydrodynamics | 54 | 23 | 62 | 87.10 | 37.10 |
| Yatch hydrodynamics | 56 | 29 | 62 | 90.32 | 46.77 |
| Yatch hydrodynamics | 46 | 16 | 62 | 74.19 | 25.81 |
| Computer Hardware | 34 | 34 | 42 | 80.95 | 80.95 |
| Computer Hardware | 37 | 42 | 42 | 88.10 | 100 |
| Computer Hardware | 35 | 23 | 42 | 83.33 | 54.76 |
| Auto | 49 | 49 | 79 | 62.02 | 62.02 |
| Auto | 55 | 45 | 79 | 69.62 | 56.96 |
| Auto | 38 | 40 | 79 | 48.10 | 50.63 |

Table 3. Comparison of local predictions on test data set using decision trees and multi-linear regression.

### 7. Future Work

The current work is being done to explain support vector regression model. This work can be extended to explanation of random forest model or any other black-box model.